\documentclass[preprint,12pt]{elsarticle}

\usepackage{graphicx}

\usepackage{amssymb}
\usepackage{amsthm}
\usepackage{graphicx,amsmath,amsfonts,amscd,amssymb,bm,cite,epsfig,epsf,url,color,algorithm,algorithmic,multirow}

\newcommand{\Ss}{{\mathcal S}}
\newcommand{\R}{\mathbb{R}}

\journal{Image and Vision Computing}

\begin{document}

\begin{frontmatter}



\title{Iterative Grassmannian Optimization for Robust Image Alignment}


\author[nuist]{Jun He \corref{cor1}} 
\ead{jhe@nuist.edu.cn}
\ead[url]{http://sites.google.com/site/hejunzz/}
 \cortext[cor1]{Corresponding author. Tel. +86 13913873052}

\author[nuist]{Dejiao Zhang}
\ead{dejiaozhang@gmail.com}
\author[umich]{Laura Balzano}
\ead{girasole@umich.edu}
\author[nuist]{Tao Tao}
 \address[nuist]{School of Electronic and Information Engineering, Nanjing University of Information Science and Technology, Nanjing, 210044, China}
 \address[umich]{Department of Electrical Engineering and Computer Science, University of Michigan, Ann Arbor, USA}


\begin{abstract}
Robust high-dimensional data processing has witnessed an exciting development in recent years. Theoretical results have shown that it is possible using convex programming to optimize data fit to a low-rank component plus a sparse outlier component. This problem is also known as Robust PCA, and it has found application in many areas of computer vision. In image and video processing and face recognition, the opportunity to process massive image databases is emerging as people upload photo and video data online in unprecedented volumes. However, data quality and consistency is not controlled in any way, and the massiveness of the data poses a serious computational challenge. In this paper we present t-GRASTA, or ``Transformed GRASTA (Grassmannian Robust Adaptive Subspace Tracking Algorithm)''. t-GRASTA iteratively performs incremental gradient descent constrained to the Grassmann manifold of subspaces in order to simultaneously estimate three components of a decomposition of a collection of images: a low-rank subspace, a sparse part of occlusions and foreground objects, and a transformation such as rotation or translation of the image. We show that t-GRASTA is $4\times$ faster than state-of-the-art algorithms, has half the memory requirement, and can achieve alignment for face images as well as jittered camera surveillance images.
\end{abstract}


\end{frontmatter}


\section{INTRODUCTION}
\label{sec:introduction}

With the explosion of image and video capture, both for surveillance and personal enjoyment, and the ease of putting these data online, we 
are seeing photo databases grow at unprecedented rates. On record we know that in July 2010,
Facebook had 100 million photo uploads per day \citep{Facebook} and Instagram had a database of 400 million
photos as of the end of 2011, with 60 uploads per second \citep{Instagram}; since then both of these databases have certainly grown immensely. In 2010, there were an estimated minimum 10,000 surveillance cameras in the city of Chicago and in 2002 an estimated 500,000 in London~\citep{chicagocams10, CCTV2002}.

These
enormous collections pose both an opportunity and a challenge for image
processing and face recognition: The opportunity is that with so much
data, it should be possible to assist users in tagging photos,
searching the image database, and detecting unusual activity or anomalies. The challenge is that the data are
not controlled in any way so as to ensure data quality and consistency across photos,
and the massiveness of the data poses a serious computational
challenge.

\begin{figure}[!htb]
	\begin{center}
		\includegraphics[width=0.60\textwidth]{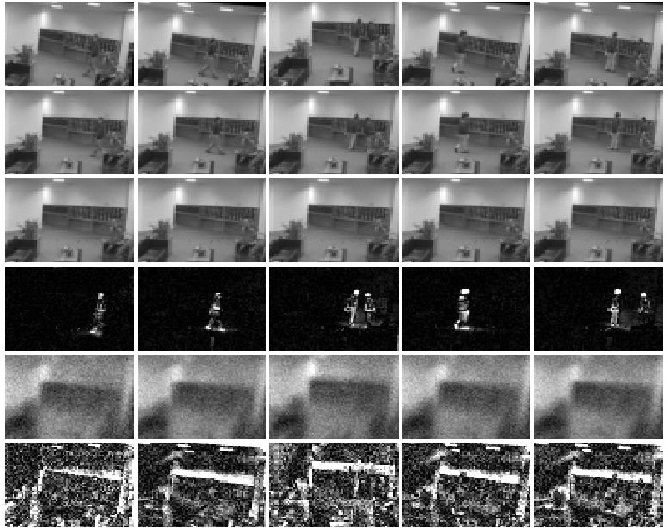} 
		\caption{  Video background and foreground separation by t-GRASTA despite camera jitter. 
		 $1^{st}$ row: misaligned video frames by simulating  camera jitters;  $2^{nd}$ row: images aligned by t-GRASTA; $3^{rd}$ row: background recovered by t-GRASTA; $4^{th}$ row: foreground separated by t-GRASTA; $5^{th}$ row: background recovered by GRASTA; $6^{th}$ row: foreground separated by GRASTA.}
		\label{fig:lobby_plot}
	\end{center}	
\end{figure}

In video surveillance, many recently proposed algorithms model the foreground and background separation problem as one of ``Robust PCA''-- decomposing the scene as the sum of a low-rank matrix of background, which represents the global appearance and illumination of the scene, and a sparse matrix of moving foreground objects \citep{Candes2011RPCA, he2012cvpr, mateos10, Sivalingam11, Torre03_RPCA}. These popular algorithms and models work very well for a stationary camera. However, in the case of camera jitter, the background is no longer low-rank, and this is problematic for Robust PCA methods \citep{jodoin2008motion,puglisi2011robust, simonson2009robust}. Robustly and efficiently detecting moving objects from an unstable camera is a challenging problem, since we need to accurately estimate both the background and the  transformation of each frame. Fig.~\ref{fig:lobby_plot} shows that for a video sequence generated by a simulated unstable camera, GRASTA \citep{he2011grasta, he2012cvpr}  (Grassmannian Robust Adaptive Subspace Tracking Algorithm) fails to do the separation, but the approach we propose here, t-GRASTA, can successfully separate the background and moving objects despite camera jitter.

Further recent work has extended the Robust PCA model to that of the ``Transformed Low-Rank +
Sparse" model for face images with occlusions that have come under transformations
such as translations and rotations~\citep{Peng2010CVPR, Peng2010PAMI, Zhang2010IJCV, wagner2012PAMI}. Without the transformations, this can be posed as a convex optimization problem and therefore convex
programming methods can be used to tackle such a problem. In RASL \citep{Peng2010PAMI} (Robust Alignment by Sparse and Low-Rank decomposition), the
authors posed the problem with transformations as well, and though it
is no longer convex it can be linearized in each iteration and proven
to reach a local minimum.

Though the convex programming methods used in~\citep{Peng2010PAMI} are polynomial in the size
of the problem, that complexity can still be too demanding for very
large databases of images. We propose Transformed GRASTA, or  t-GRASTA for short,  to tackle
this optimization with an incremental or online optimization
technique.  The benefit of this approach is three-fold: First, it will
improve speeds of image alignment both in batch mode or in online mode, as we show in Section~\ref{sec:performance}. 
Second, the memory requirement is small, which makes alignment for very large databases realistic, since t-GRASTA only needs to maintain low-rank subspaces throughout the alignment process. Finally, the proposed online version of t-GRASTA allows for alignment and occlusion removal on images as they are uploaded to the database, which is especially useful in video processing scenarios.



\subsection{Robust Image Alignment}
The problem of robust image alignment arises regularly in real data, as large illumination variations and gross pixel corruptions or partial occlusions often occur, such as sunglasses or a scarf for a human subject. The classic batch image alignment approaches, such as congealing \citep{huang2007unsupervised, learned2006data}  or  least squares congealing  algorithms \citep{cox2008least, cox2009least} cannot simultaneously handle such severe conditions, causing the alignment task to fail. 

With the breakthrough of convex relaxation theory applied to decomposing matrices into a sum of low-rank and sparse matrices \citep{chandrasekaran2011rank,Candes2011RPCA}, the recently proposed algorithm ``Robust Alignment by Sparse and Low-rank decomposition,'' or RASL \citep{Peng2010PAMI}, poses the robust image alignment problem as a transformed version of Robust PCA. The transformed batch of images can be decomposed as the sum of a low-rank matrix of recovered aligned images and a sparse matrix of errors. RASL seeks the optimal domain transformations while trying to minimize the rank of the matrix of the vectorized and stacked aligned images and while keeping the gross errors sparse. While the rank minimization and $\ell^0$ minimization can be relaxed to their convex surrogates-- minimize the corresponding nuclear norm $\|  \|_*$ and $\ell^1$ norm $\|  \|_1$-- the relaxed problem \eqref{eq:RASL_eq1} is still highly non-linear due to the complicated domain transformation.

\begin{equation}\label{eq:RASL_eq1}
	\min_{A,E,\tau}  \| A\|_* + \lambda \| E \|_1  
	\quad s.t. ~D \circ \tau  = A + E
\end{equation}
Here, $D \in \R^{n \times N}$ represents the data ($n$ pixels per each of $N$ images), $A \in \R^{n \times N}$ is the low-rank component, $E\in \R^{n \times N}$ is the sparse additive component, and $\tau$ are the transformations. RASL proposes to tackle this difficult optimization problem by iteratively locally linearizing the non-linear image transformation $D \circ (\tau + \triangle \tau) \approx D \circ \tau + \sum_{i=1}^{n} {J_i \triangle \tau_i \epsilon_i^{T}}$, where $J_i$ is the Jacobian of image $i$ with respect to transformation $i$; then in each iteration the linearized problem is convex. The authors have shown that RASL works perfectly well for batch aligning the linearly correlated images despite large illumination variations and occlusions.

In order to improve the scalability of robust image alignment for massive image datasets, \citep{Wu2012CVPR} proposes an efficient ALM-based (Augmented Lagrange Multiplier-based) iterative convex optimization algorithm ORIA (Online Robust Image Alignment) for online alignment of the input images. Though the proposed approach can scale to large image datasets, it requires the subspace of the aligned images as a prior, and for this it uses RASL to train the initial aligned subspace. Once the input images cannot be well aligned by the current subspace, the authors use an heuristic method to update the basis. In contrast, with t-GRASTA we include the subspace in the cost function, and update the subspace using a gradient geodesic step on the Grassmannian, as in~\citep{he2012cvpr, balzano2010grouse}. We discuss this in more detail in the next section.

\subsection{Online Robust Subspace Learning}

Subspace learning has been an area important to signal processing for a few decades. There are many applications in which one must track signal and noise subspaces, from computer vision to communications and radar, and a survey of the related work can be found in \citep{ comon90, Edelman98}.

The GROUSE algorithm, or ``Grassmannian Rank-One Update Subspace Estimation,'' is an online subspace estimation algorithm that can track changing subspaces in the presence of Gaussian noise and missing entries~\citep{balzano2010grouse}. GROUSE was developed as an online variant of low-rank matrix completion algorithms. It uses incremental gradient methods that have been receiving extensive attention in the optimization community~\citep{Bertsekas10}. However, GROUSE is not robust to gross outliers, and the follow-up algorithm GRASTA ~\citep{he2011grasta,he2012cvpr}, can estimate a changing low-rank subspace as well as identify and subtract outliers. Still problematic is that, as we showed in Fig.~\ref{fig:lobby_plot}, even GRASTA cannot handle camera jitter. Our algorithm includes the estimation of transformations in order to align frames first before separating foreground and background.

\section{ROBUST IMAGE ALIGNMENT VIA ITERATIVE ONLINE  SUBSPACE LEARNING}

\subsection{Model}
\label{sec:model}

\subsubsection{Batch mode} 
\label{sec:model_batch}
In order to robustly align the set of linearly correlated images despite sparse outliers, we consider the following matrix factorization model \eqref{eq:tgrasta_batch_model_nonlinear} where the low-rank matrix $U$ has orthonormal columns that span the low-dimensional subspace of the well-aligned images.

\begin{eqnarray} \label{eq:tgrasta_batch_model_nonlinear}
	\min_{U,W, E,\tau} &&\| E \|_1 \\
	s.t. && D \circ \tau = UW + E  \nonumber\\
	     && U \in \mathcal{G}(d,n) \nonumber
\end{eqnarray}
We have replaced the variable $A$ with the product of two smaller matrices $UW$, and the orthonormal columns of $U \in \R^{n \times d}$ span the low-rank subspace of the images. The set of all subspaces of $\mathbb{R}^n$ of fixed dimension $d$ is called the Grassmannian, which is a compact Riemannian manifold and is denoted by $\mathcal{G}(d,n)$. In this optimization model, $U$ is constrained to the Grassmannian $\mathcal{G}(d,n)$. Though problem \eqref{eq:tgrasta_batch_model_nonlinear} can not be directly solved \citep{Peng2010PAMI} due to the nonlinearity of image transformation, if the misalignments are not too large, by locally linearly approximating the image transformation $D \circ (\tau + \triangle \tau) \approx D\circ \tau + \sum_{i=1}^{N} J_i\triangle \tau_i \epsilon_i^{T}$, the iterative model \eqref{eq:tgrasta_batch_model_iterative} can work well as a practical approach. 

\begin{eqnarray} \label{eq:tgrasta_batch_model_iterative}
	\min_{U^{k},W, E,\triangle\tau} &&\| E \|_1 \\
	s.t. && D \circ \tau^k + \sum_{i=1}^{N} J_i^{k}\triangle \tau_i \epsilon_i^{T} = U^{k}W + E  \nonumber\\
	     && U^k \in \mathcal{G}(d^k,n) \nonumber
\end{eqnarray}
At algorithm iteration $k$, $\tau^k \doteq [\tau^k_1 \vert, \ldots, \vert\tau^k_N]$ are the current estimated transformations at iteration $k$, $J_i^k$ is the Jacobian of the $i$-th image with respect to the transformation $\tau_i^k$, and $\{\epsilon_i\}$ denotes the standard basis for $\mathbb{R}^n$. Note, at different iterations the subspace may have different dimensions, i.e. $U^k$ is constrained on different Grassmannian $\mathcal{G}(d^k,n)$. 

At each iteration of the iterative model \eqref{eq:tgrasta_batch_model_iterative}, we consider this optimization problem as the subspace learning problem. That is, our goal is to robustly estimate the low-dimensional subspace $U^k$ which best represents the locally transformed images $D \circ \tau^k + \sum_{i=1}^{N} J_i^{k}\triangle \tau_i $ despite sparse outliers $E$. In order to solve this subspace learning problem both efficiently with regards to both computation and memory, we propose to learn $U^k$ at each iteration $k$ in model \eqref{eq:tgrasta_batch_model_iterative} via the online robust subspace learning approach \citep{he2012cvpr}. 

\subsubsection{Online mode}
\label{sec:model_online}
In order to perform online video processing tasks, for example video stabilization, it is desirable to design an efficient approach that can handle image misalignment frame by frame.  As in the previous discussion regarding batch mode processing, for each video frame $I$, we may model the $\ell^1$ minimization problem as follows:

\begin{eqnarray} \label{eq:tgrasta_online_model_nonlinear}
	\min_{U,w, e,\tau} &&\| e \|_1 \\
	s.t. && I \circ \tau = Uw + e  \nonumber\\
	     && U \in \mathcal{G}(d,n) \nonumber
\end{eqnarray}
Note that with the constraint $ I \circ \tau = Uw + e $ in the above minimization problem, we suppose for each frame the transformed image is well aligned to the low-rank subspace $U$. However, due to the nonlinear geometric transform $I \circ \tau$, directly exploiting online subspace learning techniques \citep{balzano2010grouse, he2012cvpr} is not possible. 

Here we approach this as a manifold learning problem, supposing that the low-dimensional image subspace under nonlinear transformations forms a nonlinear manifold. We propose to learn the manifold approximately using a union of subspaces model $U^\ell$, $\ell = 1, \dots, L$. The basic idea is illustrated in Fig. \ref{fig:online_thought}, and the locally linearized model for the nonlinear problem \eqref{eq:tgrasta_online_model_nonlinear} is as follows: 

\begin{figure}[!h]
	\begin{center}
		\includegraphics[width=.95\textwidth]{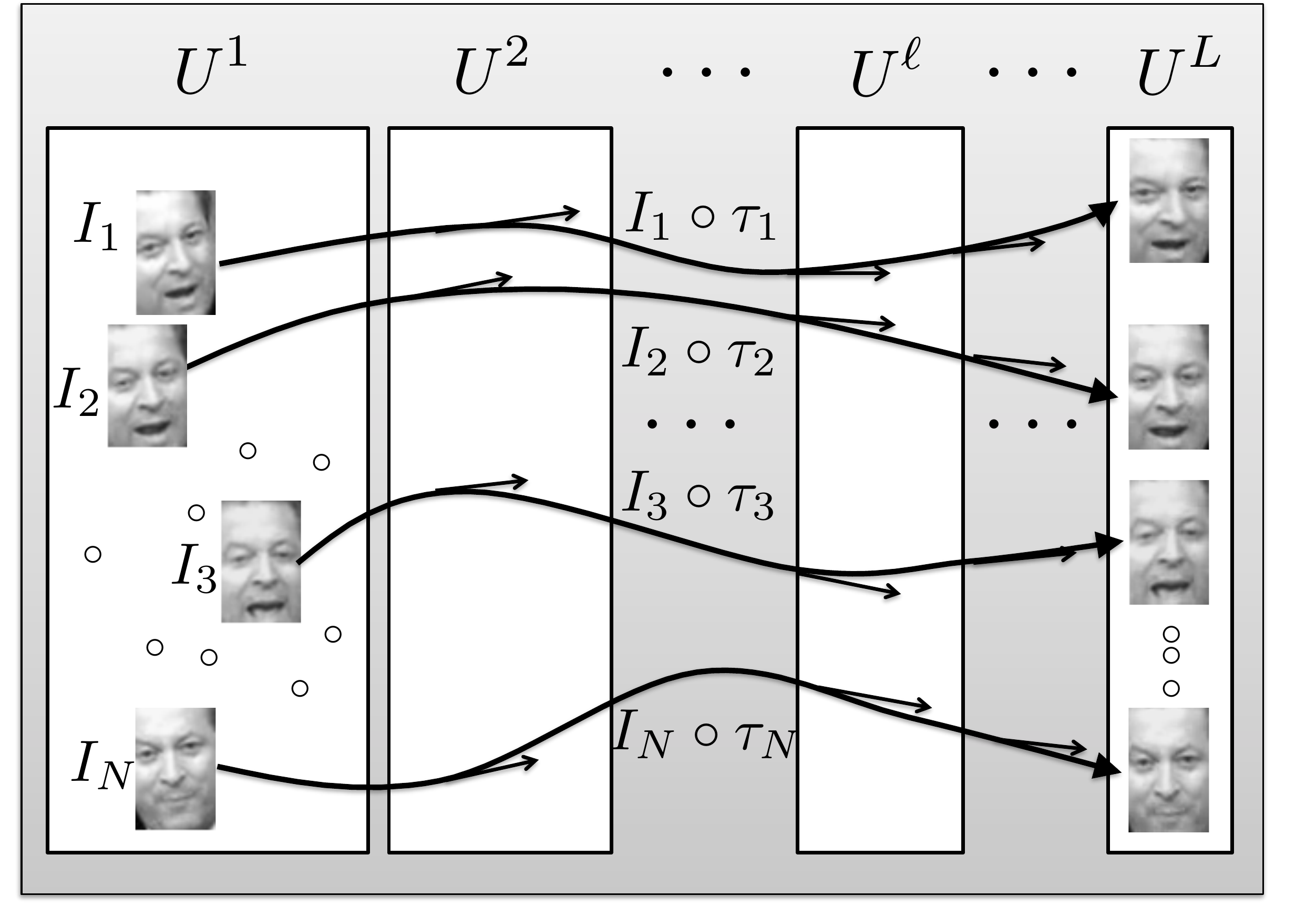}
		\caption{ The illustration of iteratively approximating the nonlinear image manifold using a union of subspaces.}
		\label{fig:online_thought}
	\end{center}	
\end{figure}

\begin{eqnarray} \label{eq:tgrasta_onine_linearized}
	\min_{w, e, \triangle{\tau}} &&\| e \|_1  \\
	 s.t. && I \circ \tau^\ell  + J^\ell \triangle{\tau}= U^\ell w + e \;.  \nonumber \\
	 && U^\ell \in \mathcal{G}(d^\ell,n) \nonumber
\end{eqnarray}

Intuitively, from Fig. \ref{fig:online_thought}, it is reasonable to think that the initial misaligned image sequence should be high rank; then after iteratively approximating the nonlinear transform with a locally linear approximation, the rank of the new subspaces $U^\ell$, $\ell = 1, \dots, L$, should be decreasing as the images become more and more aligned. Then for each misaligned image $I$ and the unknown transformation $\tau$, we iteratively update the union of subspaces $U^\ell$, $\ell = 1, \dots, L$, and estimate the transformation $\tau$. Details of the online mode of t-GRASTA will be discussed in Section \ref{sec:algos_online}

The use of a union of subspaces $U^\ell$, $\ell = 1, \dots, L$, to approximate the nonlinear manifold is a crucial innovation for this fully online model. Though we use the symbols $U^k$ and $U^\ell$ in both the batch mode and the online mode, they have two different interpretations. For batch mode, $U^k$ is the iteratively learned aligned subspace in each iteration; while for online mode, $U^\ell$, $\ell = 1, \dots, L$, is a collection of subspaces which are used for approximating the nonlinear transform, and they are updated iteratively for each video frame. 

\subsection{ADMM Solver for the Locally Linearized Problem} 
\label{sec:admm}
Whether operating in batch mode or online mode, the key problem is how to quantify the subspace error robustly for the locally linearized problem. Considering batch mode, at iteration $k$, given the $i$-th image $I_i$, its estimate of transformation $\tau_i^k$, the Jacobian $J_i^k$, and the current estimate of $U^k_t$, we use the $\ell^1$ norm as follows:

\begin{equation} \label{eq:tgrasta_lossfunc}
	F(S; t,k) = \min_{w,\triangle{\tau}} \|  U_t^k w - (I_i \circ \tau_i^k + J_i^k \triangle{\tau})\|_1
\end{equation}

With $U_t^k$ known  (or estimated, but fixed), this $\ell^1$ minimization problem is a variation of the least absolute deviations problem, which can be solved efficiently by ADMM (Alternating Direction Method of Multipliers) \citep{boyd2010distributed}. We rewrite the right hand of \eqref{eq:tgrasta_lossfunc} as the equivalent constrained problem by introducing a sparse outlier vector $e$:

\begin{eqnarray} \label{eq:tgrasta_linear}
	\min_{w, e, \triangle{\tau}} &&\| e \|_1  \\
	 s.t. && I_i \circ \tau_i^k  + J_i^k \triangle{\tau}= U_t^k w + e \;.  \nonumber
\end{eqnarray}

The augmented Lagrangian of problem \eqref{eq:tgrasta_linear} is 
 
\begin{eqnarray} \label{tgrasta_lagrangian}
	\mathcal{L}(U_t^k, w, e, \triangle{\tau}, \lambda) = \| e \|_1 &+& \lambda^{T}h(w, e, \triangle{\tau}) \nonumber \\
				&+& \frac{\mu}{2} \| h(w, e, \triangle{\tau}) \|_{2}^2
\end{eqnarray}
where $h(w, e, \triangle{\tau}) = U_t^k w + e - I_i \circ \tau_i^k - J_i^k \triangle{\tau}$, and $\lambda \in \mathbb{R}^n$ is the Lagrange multiplier or dual vector.

Given the current estimated subspace $U_t^k$, transformation parameter $\tau_i^k$, and the Jacobian matrix $J_i^k$ with respect to the $i$-th image $I_i$, the optimal $(w^*, e^*, \triangle{\tau}^*, \lambda^*)$ can be computed by the ADMM approach as follows:

\begin{equation} \label{eq:delta_tau_admm}
\left\{\begin{array}{l}
	\triangle\tau^{p+1} = (J_i^k {J_i^k}^T)^{-1} {J_i^k}^T(U_t^k w^{p} + e^{p} - I_i \circ \tau_i^k + \frac{1}{\mu}\lambda^{p}) \\
	w^{p+1} = (U_t^k {U_t^k}^T)^{-1} {U_t^k}^T( I_i \circ \tau_i^k + J_i^k\triangle\tau^{p+1} - e^{p} -  \frac{1}{\mu^p}\lambda^{p}) \\
	e^{p+1} = \textsf{S}_{\frac{1}{\mu}} (I_i \circ \tau_i^k  + J_i^k\triangle\tau^{p+1}  - U_t^k w^{p+1} - \frac{1}{\mu^p}\lambda^{p}) \\
	\lambda^{p+1} = \lambda^{p} + \mu^{p} h(w^{p+1}, e^{p+1}, \triangle\tau^{p+1})		\\
	\mu^{p+1} = \rho \mu^p
\end{array} \right.
\end{equation} where $\textsf{S}_{\frac{1}{\mu}} $ is the elementwise soft thresholding operator \citep{boyd2004convex}, and $\rho > 1$ is the ADMM penalty constant enforcing $\{\mu^p\}$ to be a monotonically increasing positive sequence. The iteration \eqref{eq:delta_tau_admm} indeed converges to the optimal solution of the problem \eqref{eq:tgrasta_linear} \citep{bertsekasnonlinear}. We summarize this ADMM solver as Algorithm \ref{alg:delta_tau} in Section \ref{sec:algos}.


\subsection{Subspace Update }
\label{sec:subspace_update}

Whether identifying the best ${U^k}^*$ in the batch mode \eqref{eq:tgrasta_batch_model_iterative} or estimating the union of subspaces $U^\ell$, $\ell = 1, \dots, L$, in the online mode \eqref{eq:tgrasta_onine_linearized}, optimizing the orthonormal matrix $U$ along the geodesic of Grassmannian is our key technique. For clarity of exposition in this section, we remove the superscript $k$ or $\ell$ from $U$, as the core gradient step along the geodesic of the Grassmannian for both batch mode and online mode is the same. We seek a sequence $\{U_t\} \in \mathcal{G}(d,n)$ such that $U_t\longrightarrow {U}^*$ (as $t\rightarrow \infty$).  We now face the choice of an effective subspace loss function. Regarding $U$ as the variable,  the loss function \eqref{eq:tgrasta_lossfunc} is not differentiable everywhere. Therefore, we choose to instead use the augmented Lagrangian \eqref{tgrasta_lagrangian} as the subspace loss function once we have estimated $(w^*, e^*, \triangle{\tau}^*, \lambda^*)$ by ADMM \eqref{eq:delta_tau_admm} from the previous $U_t$ \citep{he2011grasta, he2012cvpr}.

In order to take a gradient step along the geodesic of the Grassmannian, according to \citep{Edelman98}, we first need to derive the gradient formula of the real-valued loss function \eqref{tgrasta_lagrangian} $\mathcal{L}: \mathcal{G}(d,n) \rightarrow \mathbb{R}$. The gradient $\triangledown{\mathcal{L}}$ can be determined from the derivative of ${\mathcal{L}}$ with respect to the components of $U$:

\begin{equation} \label{eq:derivative}
	\frac{d \mathcal{L}}{d U} = \left( \lambda^* +  \mu h({w^*}, {e^*}, {\triangle\tau^*})  \right) {w^*}^T
\end{equation}
Then the gradient is $\triangledown{\mathcal{L}} = (I - U{U}^T)\frac{d \mathcal{L}}{d U}$ \citep{Edelman98}. From Step \ref{step:grad} of Algorithm \ref{alg:tgrasta}, we have that  $\triangledown{\mathcal{L}} = \Gamma {w^*}^T$ (see the definition of $\Gamma$ in Alg. \ref{alg:tgrasta}). It is easy to verify that $\triangledown{\mathcal{L}} $ is rank one since $\Gamma$ is a $n \times 1$ vector and $w^*$ is a $d \times 1$ weight vector. The following derivation of geodesic gradient step is similar to GROUSE \citep{balzano2010grouse} and GRASTA  \citep{he2011grasta, he2012cvpr}. We rewrite the important steps of the derivation here for completeness. 

The sole non-zero singular value is $\sigma = \| \Gamma  \| \|  {w^*}\|$, and the corresponding left and right singular vectors are $\frac{\Gamma}{\| \Gamma \|}$ and $\frac{w^*}{\|w^* \|}$ respectively. Then we can write the SVD of the gradient explicitly by adding the orthonormal set $x_2, \ldots, x_d$ orthogonal to $\Gamma$ as left singular vectors and the orthonormal set $y_2, \ldots, y_d$ orthogonal to $w^*$ as right singular vectors as follows:

\vspace{-4mm}
\begin{eqnarray*} \label{eq:svd_grad}
	 \triangledown{\mathcal{L}}  = && \left[ \frac{\Gamma}{\| \Gamma \|} ~~ x_2~~\ldots ~~ x_d \right] \times \text{diag}(\sigma, 0, \ldots, 0) \\&&\times \left[ \frac{w^*}{\|w^* \|} ~~ y_2 ~~\ldots ~~ y_d \right]^T\;.
\end{eqnarray*}
Finally, following Equation (2.65) in~\citep{Edelman98}, a geodesic gradient step of length $\eta$  in the direction $- \triangledown{\mathcal{L}}$ is given by 

\vspace{-4mm}
\begin{eqnarray} \label{eq:tgrasta_geodesic}
	U(\eta) = U &+& (\cos(\eta \sigma)-1) \frac{Uw_t^*}{\|w_t^*\|}\frac{ {w_t^*}^T}{\|w_t^*\|} \nonumber \\ & -& \sin(\eta \sigma)\frac{\Gamma}{\|\Gamma\|} \frac{ {w_t^*}^T}{\|w_t^*\|} \label{eq:gradient_step} \;.
\end{eqnarray}

\subsection{Algorithms}
\label{sec:algos}

\subsubsection{Batch Mode}
From the discussion of of Sections \ref{sec:admm} and \ref{sec:subspace_update}, given the batch of unaligned images $D$, their estimate of  transformation $\tau^k$ and their Jacobian $J^k$ at iteration $k$, we can robustly identify the subspace ${U^k}^*$ by incrementally updating $U^k_t$ along the geodesic of Grassmannian $\mathcal{G}(d^k,n)$ \eqref{eq:tgrasta_geodesic}.  When $U^k_t\longrightarrow {U^k}^*$ (as $t\rightarrow \infty$), the estimate of $\triangle \tau_i$ for each initially aligned image $I_i \circ \tau_i^k$ also approaches its optimal value $\triangle \tau_i ^ *$. Once the subspace $U^k$ is accurately learned, we will update the estimate of the transformation for each image using $\tau_i^{k+1} = \tau_i^k + \triangle \tau_i ^*$. Then in the next iteration, the new subspace $U^{k+1}$ can also be learned from $D \circ \tau^{k+1}$, and the algorithm iterates 
until we reach the stopping criterion, e.g. if $\frac{\|\triangle \tau  \|_2}{ \| \tau^k \|_2} < \epsilon$ or we reach the maximum iteration $K$.

We summarize our algorithms as follows. Algorithm \ref{alg:tgrasta}  is the batch image alignment approach via iterative online robust subspace learning.  For Step \ref{step:step_size}, there are many ways to pick the step-size. For some examples, you may consider the diminishing and constant step-sizes adopted in GROUSE \citep{balzano2010grouse}, or the multi-level adaptive step-size used for fast convergence in GRASTA~\citep{he2011grasta}.

\begin{algorithm}
\caption{Transformed GRASTA - batch mode}
\label{alg:tgrasta}
\textbf{Require}: An initial $n\times d^0$ orthogonal matrices $U^{0}$. A sequence of unaligned images $I_i$ and the corresponding initial transformation parameters $\tau_i^0$, $i = 1, \ldots, N$. The maximum iteration $K$.

\textbf{Return}: The estimated well-aligned subspace ${U^k}^*$ for the well-aligned images. The transformation parameters  $\tau_i^k$ for each well-aligned image.

\begin{algorithmic}[1]
\WHILE{not converged \AND $k < K$}
\STATE Update the Jacobian matrix of each image :\\
$$J_i^k = \frac{\partial{(I_i\circ \zeta)}}{\partial\zeta} \vert _{\zeta = \tau_i^k} \quad (i = 1 \ldots N)$$
\STATE Update the wrapped and normalized images:\\
$$  I_i \circ \tau_i^k  = \frac{ vec(I_i \circ \tau_i^k)}{\| vec(I_i \circ \tau_i^k) \|_2}$$

\FOR {$j=1 \to N, \ldots, ~ until ~ converged$}
\STATE Estimate the weight vector ${w_j^k}$, the sparse outliers ${e_j^k}$, the locally linearized transformation parameters ${\triangle\tau_j^k}$, and the dual vector ${\lambda_j^k}$ via the ADMM algorithm \ref{alg:delta_tau} from $  I_i \circ \tau_i^k $ , $J_i^k$, and the current estimated subspace $U_t^k$
$$ ({w_j^k}, {e_j^k}, {\triangle\tau_j^k}, {\lambda_j^k}) = \arg\min_{w,e,\triangle\tau,\lambda}\mathcal{L}(U_{t}^k, w, e, \lambda)  \label{step:admm}$$

\STATE Compute the gradient $\triangledown{\mathcal{L}} $ as follows: \label{step:grad}\\
\quad $\Gamma_1 = {\lambda_j^k} + \mu h({w_j^k}, {e_j^k}, {\triangle\tau_j^k})$,  \\
\quad $\Gamma = (I - U_t^k {U_t^k}^T)\Gamma_1 $, 
 $\qquad$ $\triangledown{\mathcal{L}} = \Gamma {w_j^k}^T$
 
\STATE Compute step-size $\eta_t$. \label{step:step_size}
 \STATE Update subspace: \\$U_{t+1}^k =  U_t^k + \left((\cos(\eta_t \sigma)-1)U_t\frac{{w_j^k}}{\|{w_j^k}\|} \right.$ \\ $\left. - \sin(\eta_t \sigma)\frac{\Gamma}{\|\Gamma\|}\right) \frac{ {w_j^k}^T}{\|{w_j^k}\|}$,
\hspace{.1in} where $\sigma = \|\Gamma\| \|{w_j^k}\|$ \;. 
\ENDFOR
\STATE Update the transformation parameters: \\$$\tau_i^{k+1} = \tau_i^k + \triangle\tau_i^k,\quad (i = 1 \ldots N)$$
 \ENDWHILE
\end{algorithmic}

\end{algorithm}

Algorithm \ref{alg:delta_tau} is the ADMM solver for the locally linearized problem \eqref{eq:tgrasta_linear}. From our extensive experiments, if we set the ADMM penalty parameter $\rho = 2$ and the tolerance $\epsilon^{tol} = 10^{-7}$, Algorithm \ref{alg:delta_tau} has always converged in fewer than $20$ iterations.

\begin{algorithm}
\caption{ADMM Solver for  the Locally Linearized Problem \eqref{eq:tgrasta_linear}}
\label{alg:delta_tau}
\textbf{Require}:  An $n \times d$ orthogonal matrix $U$, a wrapped and normalized image $I \circ \tau \in \R^{n}$ , the corresponding Jacobian matrix $J$, and a structure OPTS which holds four parameters for ADMM: ADMM penalty constant $\rho$, the tolerance $\epsilon^{tol}$, and ADMM maximum iteration $K$.

\textbf{Return}: weight vector $w^*\in \R^d$; sparse outliers $e^*\in \R^n$; locally linearized transformation parameters ${\triangle\tau^*}$ ; and dual vector $\lambda^*\in \R^n$.

\begin{algorithmic}[1]
\STATE Initialize $w, e, \triangle\tau, \lambda, and ~\mu$: $e^1 = 0$,$w^1 = 0$,$\triangle\tau^1 = 0$, $\lambda^1 = 0$, $\mu = 1$
\STATE Cache $P = (U^T U)^{-1}U^T$ and $F = (J^T J)^{-1}J^T$
\FOR{$k = 1 \to K$ } 
\STATE Update ${\triangle\tau}$:  $\triangle\tau^{k+1} = F(Uw^k + e^k - I \circ \tau + \frac{1}{\mu}\lambda^k)$
\STATE Update weights: $w^{k+1} = P (I \circ \tau  + J\triangle\tau^{k+1} - e^k - \frac{1}{\mu}\lambda^k)$ 
\STATE Update sparse outliers: \\$e^{k+1} = \textsf{S}_{\frac{1}{\mu}} (I \circ \tau  + J\triangle\tau^{k+1}  - Uw^{k+1} - \frac{1}{\mu}\lambda^k) $
\STATE Update dual: $\lambda^{k+1} = \lambda^k + \mu h(w^{k+1}, e^{k+1}, \triangle\tau^{k+1}) $
\STATE Update $\mu$: $\mu =  \rho \mu$
\IF {$\|  h(w^{k+1}, e^{k+1}, \triangle\tau^{k+1}) \|_2 \leq \epsilon^{tol}$}
\STATE Converge and break the loop.
\ENDIF
\ENDFOR
\STATE $w^* = w^{k+1}$, $e^*=e^{k+1}$, ${\triangle\tau^*} = \triangle\tau^{k+1}$ , $\lambda^* = y^{k+1}$
\end{algorithmic}

\end{algorithm}

\subsubsection{Online Mode}
\label{sec:algos_online}

\begin{figure}[!h]
	\begin{center}
		\includegraphics[width=.95\textwidth]{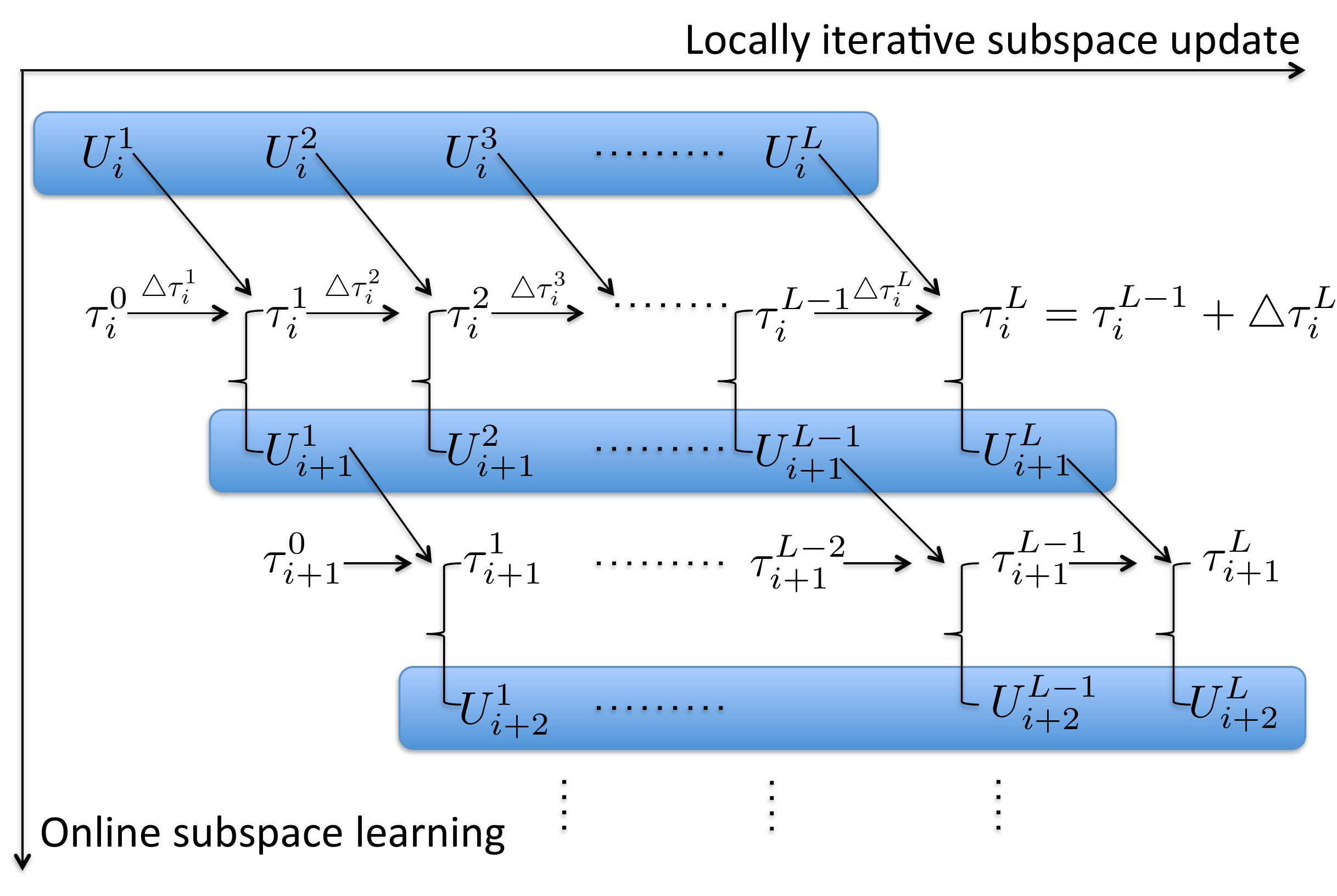}
		\caption{ The diagram of the \textit{fully online mode} of t-GRASTA.}
		\label{fig:online_diagram}
	\end{center}	
\end{figure}

In Section \ref{sec:model_online}, we propose to tackle the difficult nonlinear online subspace learning problem by iteratively learning online a union of subspaces  $U^\ell$, $\ell = 1, \dots, L$. For a sequence of video frames $I_i, i=1, \dots, N$, the union of subspaces  $U^\ell$ are updated iteratively as illustrated in Fig. \ref{fig:online_diagram}. 

Specifically, at $i$-th frame  $I_i$, for the locally approximated subspace $U_i^1$ at the first iteration, given the initial roughly estimated transformation $\tau_i^0$, the ADMM solver Algorithm \ref{alg:delta_tau} gives us  the  locally estimated $\triangle \tau_i^1$, and the updated subspace $U_{i+1}^1$ is obtained by taking a gradient step along the geodesic of the Grassmannian $\mathcal{G}(d^1,n)$ as discussed in Section \ref{sec:subspace_update}. The transformation $\tau_i^1$ of the next iteration is updated by $\tau_i^1 = \tau_i^0 + \triangle \tau_i^1$. Then for the next locally approximated subspace $U_i^2$, we also estimate $\triangle \tau_i^2$ and update the subspace along the geodesic of the Grassmannian $\mathcal{G}(d^2,n)$ to $U_{i+1}^2$. Repeatedly, we will update $U_i^\ell$ in the same way to get $U_{i+1}^\ell$ and the new transformation $\tau_i^\ell = \tau_i^{\ell -1} + \triangle \tau_i^\ell$ . After completing the update for all $L$ subspaces, the union of subspaces $U_{i+1}^\ell (\ell = 1, \dots, L)$ will be used for approximating the nonlinear transform of the next video frame $I_{i+1}$.

We summarize the above statements as Algorithm \ref{alg:tgrasta-online}, and we call this approach the \textit{fully online mode} of t-GRASTA. 

\begin{algorithm}
\caption{Transformed GRASTA - \textit{Fully Online Mode}}
\label{alg:tgrasta-online}
\textbf{Require}: The initial $L$ $n\times d^\ell$ orthonormal matrices $U^{\ell}$ spanning the corresponding subspace $\Ss^\ell$, $\ell= 1,...,L$. A sequence of unaligned images $I_i$ and the corresponding initial transformation parameters $\tau_i^0$, $i = 1, \ldots, N$. 

\textbf{Return}: The estimated iteratively approximated subspaces $U_i^\ell$, $\ell = 1,\ldots,L$, after processing image $I_i$. The transformation parameters  $\tau_i^L$ for each well-aligned image.

\begin{algorithmic}[1]
\FOR {unaligned image $I_i, i=1,\ldots,N$}
\FOR {the iterative approximated subspace $U^\ell, \ell=1,\ldots, L$}
\STATE Update the Jacobian matrix of  image $I_i$:\\
$$J_i^\ell = \frac{\partial{(I_i\circ \zeta)}}{\partial\zeta} \vert _{\zeta = \tau_i^\ell} $$
\STATE Update the wrapped and normalized images:\\
$$  I_i \circ \tau_i^\ell  = \frac{ vec(I_i \circ \tau_i^\ell)}{\| vec(I_i \circ \tau_i^\ell) \|_2}$$

\STATE Estimate the weight vector ${w_i^\ell}$, the sparse outliers ${e_i^\ell}$, the locally linearized transformation parameters ${\triangle\tau_i^\ell}$, and the dual vector ${\lambda_i^\ell}$ via the ADMM algorithm \ref{alg:delta_tau} from $  I_i \circ \tau_i^\ell $ , $J_i^\ell$, and the current estimated subspace $U_i^\ell$
$$ ({w_i^\ell}, {e_i^\ell}, {\triangle\tau_i^\ell}, {\lambda_i^\ell}) = \arg\min_{w,e,\triangle\tau,\lambda}\mathcal{L}(U_{i}^\ell, w, e, \lambda)  \label{step:admm}$$

\STATE Compute the gradient $\triangledown{\mathcal{L}} $ as follows: \label{step:grad}\\
\quad $\Gamma_1 = {\lambda_i^\ell} + \mu h({w_i^\ell}, {e_i^\ell}, {\triangle\tau_i^\ell})$,  \\
\quad $\Gamma = (I - U_t^\ell {U_t^\ell}^T)\Gamma_1 $, 
 $\qquad$ $\triangledown{\mathcal{L}} = \Gamma {w_i^\ell}^T$
 
\STATE Compute step-size $\eta_i^\ell$. \label{step:step_size}
 \STATE Update subspace: \\$U_{i+1}^\ell =  U_i^\ell + \left((\cos(\eta_i^\ell \sigma)-1)U_t\frac{{w_i^\ell}}{\|{w_i^\ell}\|} \right.$  $\left. - \sin(\eta_i^\ell \sigma)\frac{\Gamma}{\|\Gamma\|}\right) \frac{ {w_i^\ell}^T}{\|{w_i^\ell}\|}$,\\
\hspace{.1in} where $\sigma = \|\Gamma\| \|{w_i^\ell}\|$ \;. 
\STATE Update the transformation parameters: \\$$\tau_i^{\ell+1} = \tau_i^\ell + \triangle\tau_i^\ell$$
\ENDFOR
\ENDFOR
\end{algorithmic}

\end{algorithm}

\subsubsection{Discussion of Online Image Alignment}
If the subspace $U^k$ of the well-aligned images is known as a prior, for example if $U^k$ is trained by Algorithm \ref{alg:tgrasta} from a ``well selected'' dataset of one category, we can simply use $U^k$ to align the rest of the unaligned images of the same category. Here ``well selected" means the training dataset should cover enough of the global appearance of the object, such as different illuminations, which can be represented by the low-dimensional subspace structure. By category, we mean a particular object of interest or a particular background scene in the video surveillance data.

For massive image processing tasks, it is easy to collect such good training datasets by simply randomly sampling a small fraction of the whole image set. Once $U^k$ is learned from the training set, we can use a variation of Algorithm \ref{alg:tgrasta} to align each unaligned image $I$ without updating the subspace, since we have the assumption that the remaining images also lie in the trained subspace. We call Algorithm  \ref{alg:simple_online} the \textit{trained online mode}.

However, we note that for a very large streaming dataset such as is typical in real-time video processing, the \textit{trained online mode} may be less well-defined, as the subspace of the streaming video data may change over time. For this scenario, our \textit{fully online mode} for t-GRASTA could gradually adapt to the changing subspace and then accurately estimate the transformation $\tau$.

\subsection{Discussion of Memory Usage}
We compare the memory usage of our \textit{ fully online mode} of t-GRASTA  to that of RASL.  RASL requires storage of $A$, $E$, a Lagrange multiplier matrix $Y$, the data $D$, and $D \circ \tau$, each of which require storage of the size $nN$. To compare fairly to t-GRASTA, which assumes a $d$-dimensional model, we suppose RASL uses a thin singular value decomposition of size $d$, which requires $nd + Nd + d^2$ memory elements. Finally for the Jacobian per image, RASL needs $nNp$, and for $\tau$ RASL needs $Np$, but we will assume $p$ is a small constant independent of dimension and ignore it. Therefore RASL's total memory usage is $6nN+nd+Nd+d^2+N$.

t-GRASTA must also store the Jacobian, $\tau$, and the data as well as the data with transformation, using memory size $3nN+N$. Otherwise, t-GRASTA needs to store the union of subspaces $U^\ell$, $\ell = 1, \dots, L$ matrices of size $Lnd (L \ll N)$, and the vectors $e$, $\lambda$, $\Gamma$, and $w$ for $3n+d$ memory elements. Thus t-GRASTA's memory total is $3nN+Lnd+3n+d+N$. 

For a problem size of 100 images, each with 100$\times$100 pixels, and assuming $d=10$, $L=10$, t-GRASTA uses 66.1\% of the memory of RASL. For 10000 mega-pixel images, t-GRASTA uses 50.1\% of the memory of RASL. The scaling remains about half throughout mid-range to large problem sizes.

\begin{algorithm}
\caption{\textit{Trained Online Mode} of Image Alignment}
\label{alg:simple_online}
\textbf{Require}: A well-trained $n\times d$ orthogonal matrix $U$. An unaligned image $I$ and the corresponding initial transformation parameters $\tau^0$. The maximum iteration $K$.

\textbf{Return}: The transformation parameters  $\tau^k$ for the well-aligned image.

\begin{algorithmic}[1]
\WHILE{not converged \AND $k < K$}
\STATE Update the Jacobian matrix  :\\
$$J^k = \frac{\partial{(I\circ \zeta)}}{\partial\zeta} \vert _{\zeta = \tau^k}$$
\STATE Update the wrapped and normalized image:\\
$$  I \circ \tau^k  = \frac{ vec(I \circ \tau^k)}{\| vec(I \circ \tau^k) \|_2}$$


\STATE Estimate the weight vector ${w^k}$, the sparse outliers ${e^k}$, the locally linearized transformation parameters ${\triangle\tau^k}$, and the dual vector ${\lambda^k}$ via the ADMM algorithm \ref{alg:delta_tau} from $  I \circ \tau^k $ , $J^k$, and the well-trained subspace $U$
$$ ({w^k}, {e^k}, {\triangle\tau^k}, {\lambda^k}) = \arg\min_{w,e,\triangle\tau,\lambda}\mathcal{L}(U, w, e, \lambda)  $$
 
\STATE Update the transformation parameters: \\$$\tau^{k+1} = \tau^k + \triangle\tau^k$$
 \ENDWHILE
\end{algorithmic}

\end{algorithm}

\addtolength{\textheight}{0cm}   

\section{PERFORMANCE EVALUATION}
\label{sec:performance}
In this section, we conduct comprehensive experiments on a variety of alignment tasks to verify the efficiency and superiority of our algorithm. We first demonstrate the ability of the proposed approach to cope with occlusion and illumination variation during the alignment process. After that, we further demonstrate the robustness and generality of our approach by testing it on handwritten digits and face images taken from the Labeled Faces in the Wild database~\citep{LFW}. Finally, we apply our approach to dealing with video jitters and solving the interesting background foreground separation problem. 

\subsection{Occlusion and illumination variation}
\begin{figure}[!h]
	\begin{center}
		\includegraphics[width=.95\textwidth]{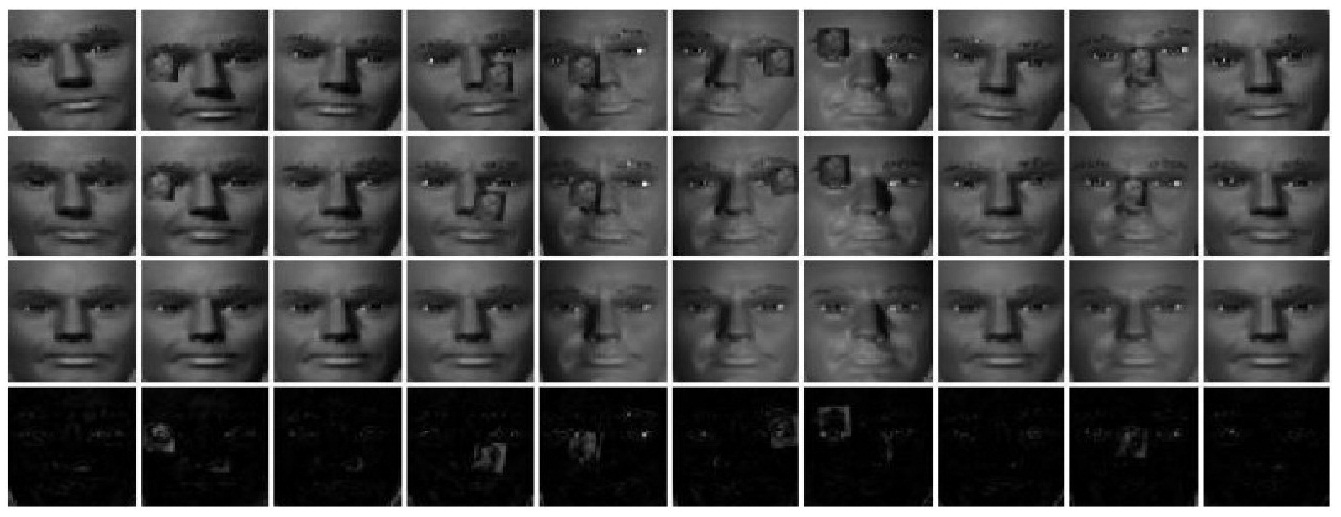}
		\caption{
		The first row shows the original misaligned images with occlusions and illumination variation; the second row shows the images aligned by t-GRASTA; the third row shows the recovered aligned images without occlusion; and the bottom row is the occlusion removed by our approach.}
		\label{fig:fig_dummy}
	\end{center}	
\end{figure}

We first test our approach on the dataset `dummy' described in \citep{Peng2010PAMI}. Here, we want to verify the ability of our approach to effectively align the images despite occlusion and illumination variation. The dataset contains 100 images of a dummy head taken under varying illumination and with artificially generated occlusions created by adding a square patch at a random location of the image. Fig.~\ref{fig:fig_dummy} shows 10 misaligned images of the dummy. We align these images by Algorithm \ref{alg:tgrasta} (the batch mode of t-GRASTA). The canonical frame is chosen to be $49 \times 49$ pixels and the subspace dimension is set to 5. Here and in the rest of our experiments, for simplicity we set $d^k$ of Algorithm \ref{alg:tgrasta} to a fixed $d$ in every iteration. The last three rows of Fig.~\ref{fig:fig_dummy} show the results of alignment, from which we can see that our approach is successful at aligning the misaligned images while removing the occlusion at the same time.

\begin{figure*}[!htb]
	\begin{center}
		\includegraphics[width=.8\textwidth]{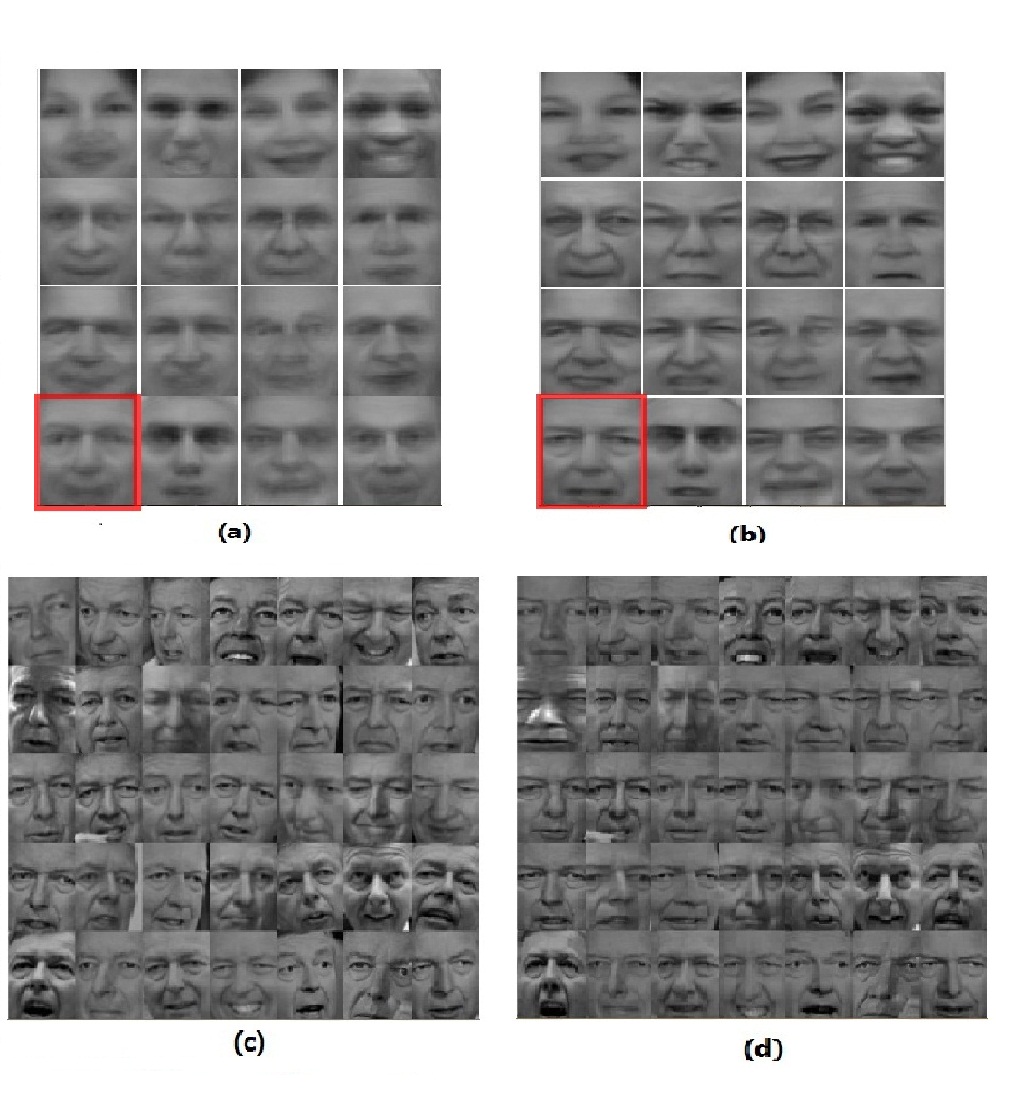}
		\caption{ (a) Average of 16 misaligned subjects randomly selected from LFW database; (b) average of each subject aligned by t-GRASTA; (c) initial images of John Ashcroft (marked by red boxs in (a) and (b)); (d) images aligned by t-GRASTA.}
		\label{fig:fig_LFW}
	\end{center}	
\end{figure*} 

\subsection{Robustness}

In order to further demonstrate the robustness of our approach, we apply it on more realistic images taken from the Labeled Faces in the Wild database~\citep{LFW}. The LFW contains more severely misaligned images, for it also includes remarkable variations in pose and expression aside from illumination and occlusion, which can be seen in Fig.~\ref{fig:fig_LFW}(c). We chose 16 subjects from LFW, each of them with 35 images. Each image is aligned to an $80 \times 60$ canonical frame using $\tau$ which are from the group of affine transformations $\mathbb{G} = Aff(2)$, as in~\citep{Peng2010PAMI}; these are translations, rotations, and scale transformations. For each subject, we set the subspace dimension = 15 and use Algorithm \ref{alg:tgrasta} to align each image. In this example, we demonstrate the robustness of our approach by comparing the average face of each subject before and after alignment, which are shown in Fig.~\ref{fig:fig_LFW}(a)-(b). We can see that the average faces after alignment are much clearer than those before alignment. Fig.~\ref{fig:fig_LFW}(c)-(d) provides more detailed information, showing the unaligned and aligned images of John Ashcroft (marked by red boxes in Fig.~\ref{fig:fig_LFW}(a)-(b)).

\subsection{Generality}

The previous experiments have demonstrated the effectiveness and robustness of t-GRASTA. Here we wish to show the generality of t-GRASTA by applying it to aligning a different type of images -- handwritten digits taken from MINST database. For this experiment, we again use Algorithm \ref{alg:tgrasta} to align 100 images of a handwritten ``3" to a $29 \times 29$ canonical frame size. We use Euclidean transformation $\mathbb{G} = E(2)$ and set the dimension of the subspace to be 5. 

\begin{figure*}[!htb]
	\begin{center}
		\includegraphics[width=.9\textwidth]{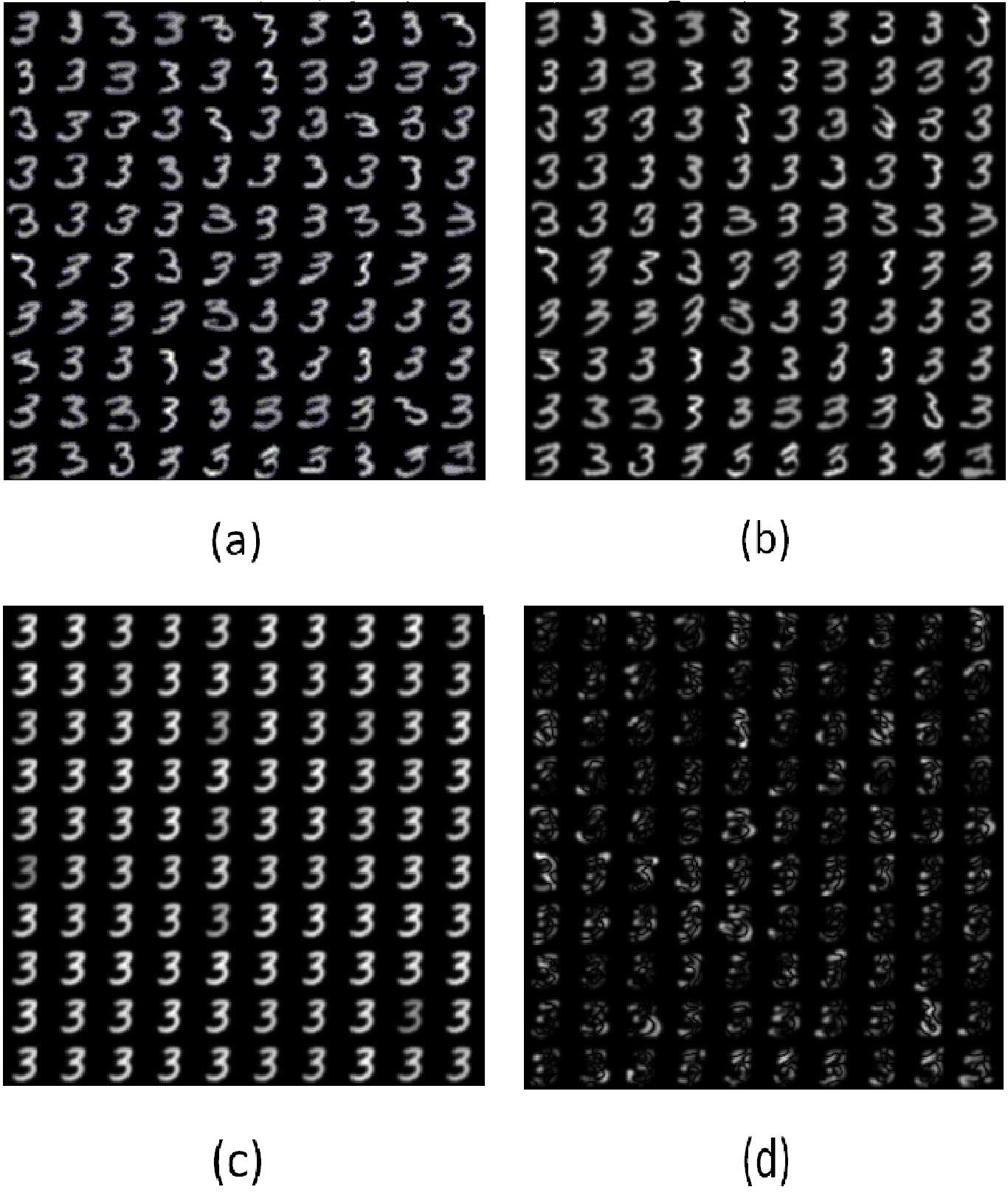}
		\caption{ (a) 100 misaligned digits; (b) digits aligned by t-GRASTA; (c) subspace representation of corresponding digits; (d) outliers.}
		\label{fig:fig_digit}
	\end{center}	
\end{figure*}

Fig.~\ref{fig:fig_digit} shows that t-GRASTA can successfully align the misaligned digits and learn the low dimensional subspace, even though the original digits have significant variation. We can see that the outliers separated by t-GRASTA are generated by variations in the digits that are not consistent with the global appearance. The outliers (d) would be even more sparse if the subspace representation in (c) were to capture more of this variation; If desired, we could achieve this tradeoff by increasing the dimension of the subspace.   

\subsection{Video Jitter}
In this section, we apply t-GRASTA to separation problems made difficult by video jitter. Here we apply both the \textit{fully online mode} Algorithm \ref{alg:tgrasta-online} and the \textit{trained online mode} Algorithm \ref{alg:simple_online} to different datasets. We show the superiority of t-GRASTA regarding both the speed and memory requirement of the algorithms. 

\subsubsection{Hall}
Here we apply t-GRASTA to the task of separating moving objects from static background in the video footage recorded by an unstable camera. We note that in \citep{he2012cvpr}, the authors simulate a virtual panning camera to show that GRASTA can quickly track sudden changes in the background subspace caused by a moving camera. Their low-rank subspace tracking model is well-defined, as the camera after panning is still stationary, and thus the recorded video frames are accurately pixelwise aligned. However, for an unstable camera, the recorded frames are no longer aligned; the background cannot be well represented by a  low-rank  subspace unless the jittered frames are first aligned. In order to show that t-GRASTA can tackle this separation task, we consider a highly jittered video sequence generated by a simulated unstable camera. To simulate the unstable camera, we randomly translate the original well-aligned video frames in x- / y- axis and rotate them in the plane.

In this experiment, we compare t-GRASTA with RASL and GRASTA. We use the first 200 frames of the ``Hall" dataset\footnote{Find these along with the videos at \url{http://perception.i2r.a-star.edu.sg/bk_model/bk_index.html}.}, each $144 \times 176$ pixels. We first perturb each frame artificially to simulate camera jitter. The rotation of each frame is random, uniformly distributed within the range of [$-\theta_0/2, \theta_0/2$], and the ranges of x- and y-translations are limited to [$-x_0 /2$,$x_0/2$] and [$-y_0 /2$,$y_0 /2$]. In this example, we set the perturbation size parameters [$x_0$,$y_0$,$\theta_0$] with the values of [ 20,20,$10^{\circ}$].

\begin{figure}[!htb]
	\begin{center}
		\includegraphics[width=0.60\textwidth]{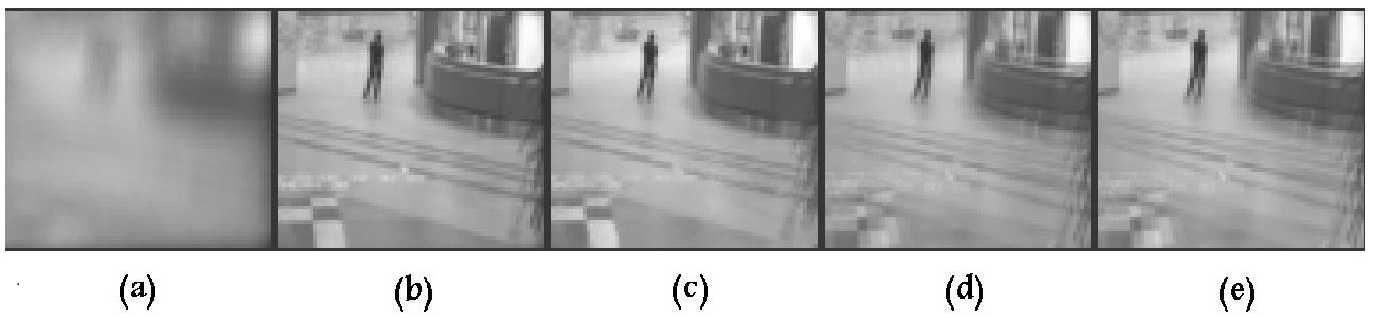}
		\caption{Comparison between t-GRASTA and RASL. (a) Average of initial misaligned images; (b) average of images aligned by t-GRASTA; (c)average of background recovered by t-GRASTA; (d) average of images aligned by RASL; (e) average of background recovered by RASL.}
		\label{fig:fig_comparison}
	\end{center}	
\end{figure}

For comparing with RASL, unlike \citep{Wu2012CVPR}, we just let RASL run its original batch model without forcing it into an online algorithm framework. The task we give to RASL and t-GRASTA is to align each frame to a $62 \times 75$ canonical frame, again using $\mathbb{G} = Aff(2)$. The dimension of the subspace in t-GRASTA is set to be 10. We first randomly select 30 frames of the total 200 frames to train the subspace by Algorithm \ref{alg:tgrasta} and then align the rest using the \textit{trained online mode}. The visual comparison between RASL and t-GRASTA are shown in Fig.~\ref{fig:fig_comparison}. Table~\ref{tbl:rasl_tgrasta_stat_err} illustrates the numerical comparison of RASL and t-GRASTA, for which we ran each algorithm 10 times to get the statistics. From Table~\ref{tbl:rasl_tgrasta_stat_err} and Fig.~\ref{fig:fig_comparison} we can see that the two algorithms achieve a very similiar effect, but t-GRASTA runs much faster than RASL: On a PC with Intel P9300 2.27GHz CPU and 2 GB of RAM, the average time for aligning a newly arrived frame is 1.1 second, while RASL needs more than 800 seconds to align the total batch of images, or 4 seconds per frame. Moreover, our approach is also superior to RASL regarding memory efficiency. These superiorities become more dramatic as one increases the size of the image database.

\begin{table}[!htpb]
\caption{\textnormal{Statistics of errors in two pixels $P_1$ and $P_2$, selected from the original video frames and traced through the jitter simulation process to the RASL and t-GRASTA output frames. Max error and mean error are calculated as the distances from the estimated $P_1$ and $P_2$ to their statistical center $E(P_1)$ and $E(P_2)$. Std are calculated as the standard deviation of four coordinate value $(X_1,Y_1)$ for $P_1$ and $(X_2,Y_2)$ for $P_2$ across all frames.}}

\scriptsize
\begin{center}
	\begin{tabular}[width=.95\textwidth]{ c | c | c | c | c | c | c  }
	\hline
		             & Max & Mean & X1 std & Y1 std & X2 std & Y2 std \\
           & error & error\\  \hline
		Initial misalignment & 11.24 & 5.07 & 3.35 & 3.01 & 3.34 & 4.17 \\
					RASL & \textbf{2.96} & 1.73 			& 0.56 				 & \textbf{0.71} & 0.90 			& 1.54\\
           t-GRASTA & 6.62 				& \textbf{0.84} & \textbf{0.48} & 1.11 			& \textbf{0.57} & \textbf{0.74}\\   \hline
	\end{tabular}
\end{center}
		\label{tbl:rasl_tgrasta_stat_err}
\end{table}

\begin{figure*}[!htpb]
	\begin{center}
		\includegraphics[width=1\textwidth]{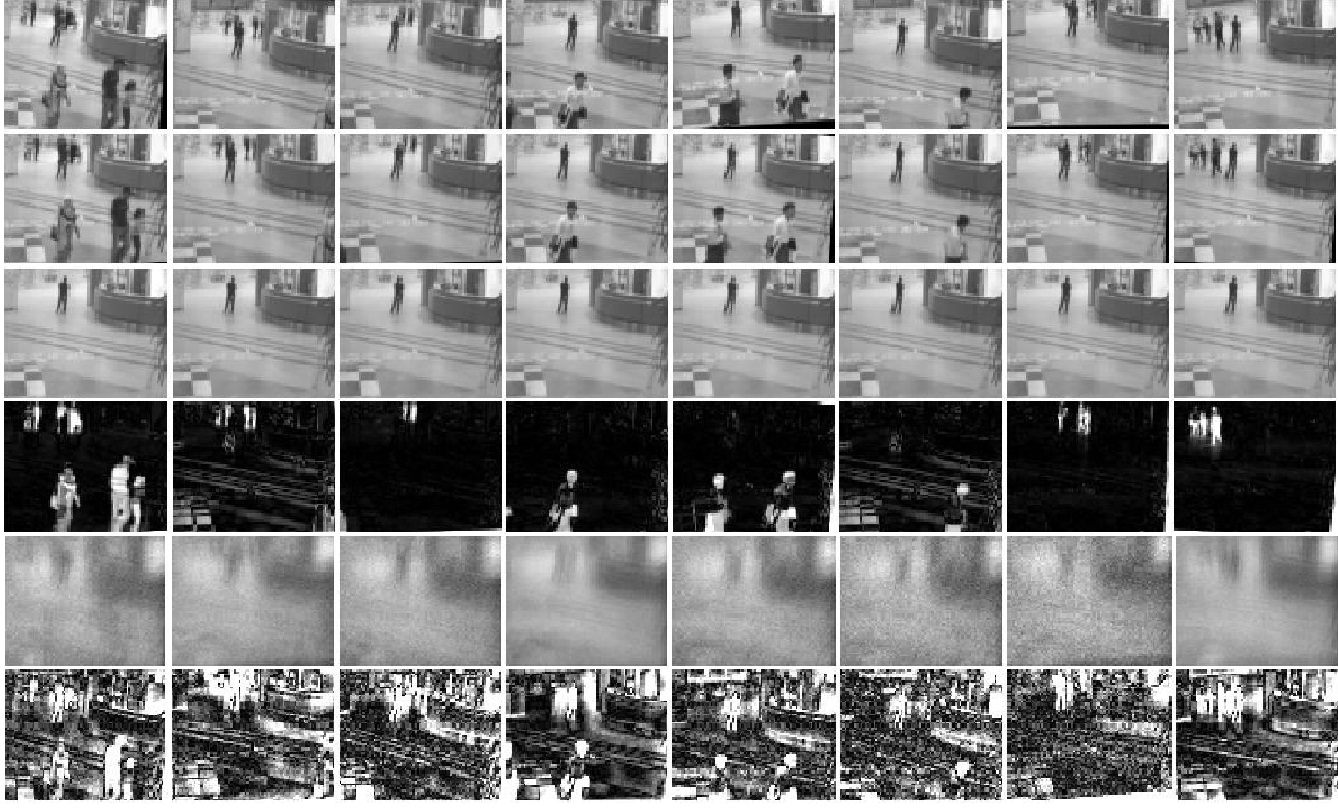}
		\caption{Video background and foreground separation with jittered video. $1^{st}$ row: 8 misaligned video frames randomly selected from artificially perturbed images;  $2^{nd}$ row: images aligned by t-GRASTA; $3^{rd}$ row: background recovered by t-GRASTA; $4^{th}$ row: foreground separated by t-GRASTA; $5^{th}$ row: background recovered by GRASTA; $6^{th}$ row: foreground separated by GRASTA.  }
		\label{fig:fig_hall}
	\end{center}	
\end{figure*}

In order to compare with GRASTA, we use 200 perturbed images to recover the background and separate the moving objects for both algorithms; Fig.~\ref{fig:fig_hall} illustrates the comparison. For both GRASTA and t-GRASTA, we set the subspace rank = 10 and randomly selected 30 images to train the subspace first. For t-GRASTA, we use the affine transformation $\mathbb{G} = Aff(2)$. From Fig.~\ref{fig:fig_hall}, we can see that our approach successfully separates the foreground and the background and simultaneously align the perturbed images. But GRASTA fails to learn a proper subspace, thus, the separation of background and foreground is poor. Although GRASTA has been demonstrated to successfully track a dynamic subspace, e.g. the panning camera, the dynamics of an unstable camera are too fast and unpredictable for the GRASTA subspace tracking model to succeed in this context without pre-alignment of the video frames.


\subsubsection{Gore}
In this example, we show the capability of t-GRASTA for video stabilization applied to the dataset ``Gore" described in \citep{Peng2010PAMI}. In \citep{Peng2010PAMI}, the original face images are obtained by a face detector, and the jitters are caused by the inherent imprecision of the detector. In contrast, for t-GRASTA, we simply crop the face from each image by a constant rectangle with size $68 \times 44$, which has the same position parameters for all frames. So in our case, the jitters are caused by the differences between the motion and pose variation of the target and the stabilization of the constant rectangle. 

\begin{figure*}[!htpb]
	\begin{center}
		\includegraphics[width=1\textwidth]{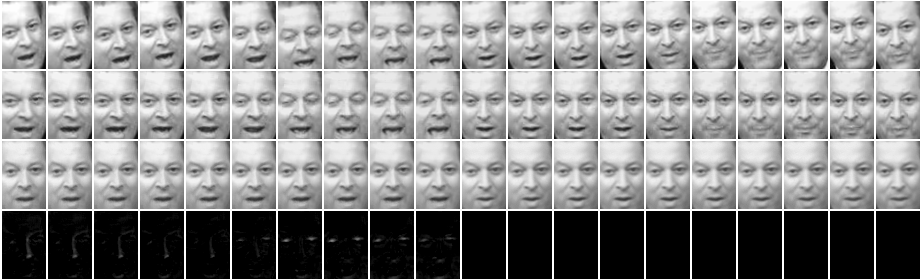}
		\caption{ The first row shows the original misaligned images; the second row shows the images aligned by t-GRASTA; the third row shows the recovered aligned images without outliers; and the bottom row shows the outliers removed by our approach.}
		\label{fig:fig_gore}
	\end{center}	
\end{figure*}

For this experiment, the dimension of the subspace is set to be 10, and we again choose the affine transformation $\mathbb{G} = Aff(2)$. We first use the Algorithm \ref{alg:tgrasta} to train an initial subspace by 20 images randomly selected from the whole set of 140 images. We then use the \textit{fully online mode} to align the rest of the images. Fig.~\ref{fig:fig_gore} show the results. t-GRASTA did well for this dataset with better speed than RASL: On a PC with Intel P9300 2.27GHz CPU and 2 GB of RAM, t-GRASTA aligned these images at 5 frames per second. This is 5 times faster than RASL and 3 times faster than ORIA as described in \citep{Wu2012CVPR}.

Although t-GRASTA was not designed as a face detector, the experimental results suggest that t-GRASTA can be transformed into a face detector, or more generally target tracker, if the variation of pose of the target is limited in a certain range (usually $45^{\circ}$). In this case, we can further improve the efficiency of t-GRASTA by choosing a tight frame for the canonical image.

\subsubsection{Sidewalk}

In the last experiment, we use misaligned frames caused by real camera jitter to test t-GRASTA. Here we align all 1200 frames of ``Sidewalk" dataset\footnote{Find it along with other datasets containing misaligned frames caused by real video jitters at \url{http://wordpress-jodoin.dmi.usherb.ca/dataset}.} to $50 \times 78$ canonical frames, again using $\mathbb{G} = Aff(2)$ and subspace dimension 5. We also use the first 20 frames to train the initial subspace using the batch mode Algorithm \ref{alg:tgrasta}, and then use the \textit{fully online mode} to align the rest of the frames.  Here we can see that aligning the total 1200 frames is a heavy task for RASL -- for our PC with Intel P9300 2.27GHz CPU and 2 GB of RAM, it was necessary to divide the dataset into four parts each containing 300 frames. We then let RASL separately run on each sub-dataset. The total time needed by RASL was around 1000 seconds for 1.2 frames per second, while t-GRASTA achieved more than 4 frames per second without partitioning the data. 

Compared to the \textit{trained online mode}, the \textit{fully online mode} can track changes of the subspace over time. This is an important asset of the \textit{fully online mode}, especially when it comes to large streaming datasets containing considerable variations. 
We see that we usually need no more than 20 frames for \textit{fully online mode} to adapt to the changes of the subspace, such as illumination changes or dynamic background caused by the motion of the subspace. Moreover, if the changes are slow, i.e the natural illumination changes from daylight or the camera moving slowly, then t-GRASTA needs no extra frames to track such changes; it incorporates such information with each iteration during the slowly changing process.
\begin{figure*}[!htpb]
	\begin{center}
		\includegraphics[width=1\textwidth]{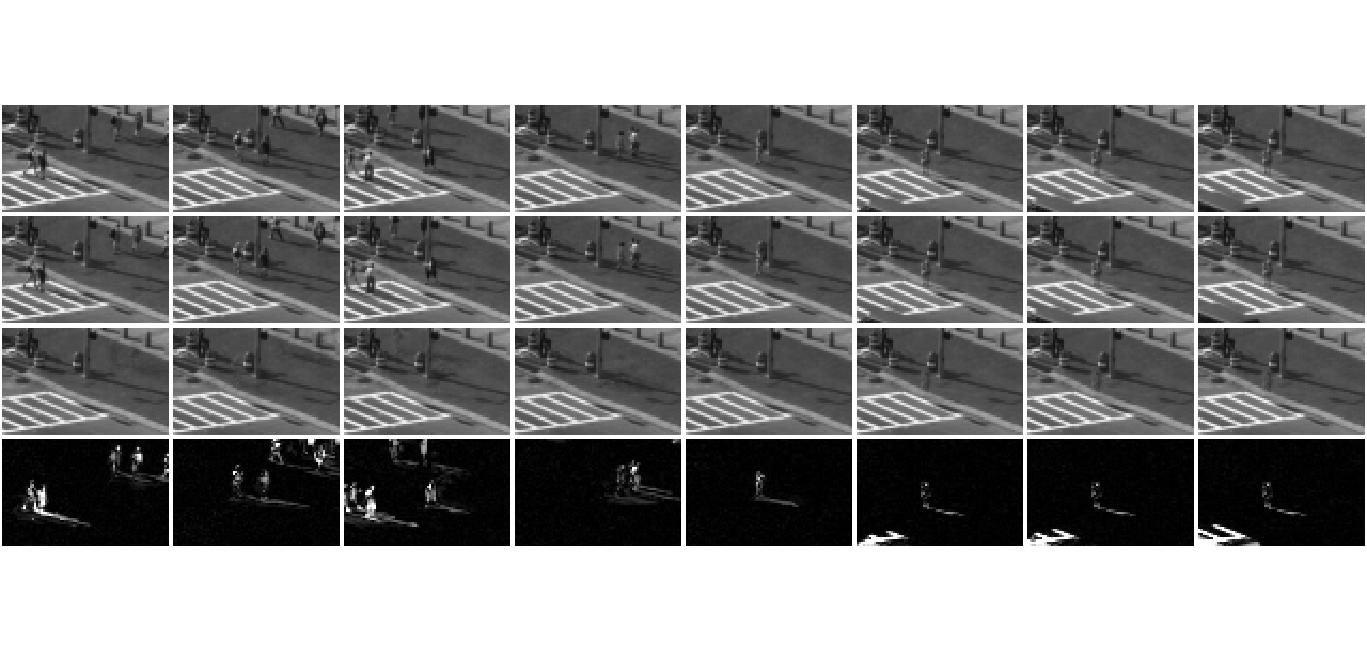}
		\caption{Video background and foreground separation with jittered video. $1^{st}$ row: 8 original misaligned video frames caused by video jitter;  $2^{nd}$ row: images aligned by t-GRASTA; $3^{rd}$ row: background recovered by t-GRASTA; $4^{th}$ row: foreground separated by t-GRASTA. }
		\label{fig:fig_sidewalk}
	\end{center}	
\end{figure*}

\section{CONCLUSIONS AND FUTURE WORK}

\subsection{Conclusions}
In this paper we have presented an iterative Grassmannian optimization approach to simultaneously identify an optimal set of image domain transformations for image alignment and the low-rank subspace matching the aligned images. These are such that the vector of each transformed image can be decomposed as the sum of a low-rank part of the recovered aligned image and a sparse part of errors. This approach can be regarded as an extension of GRASTA and RASL: We extend GRASTA to transformations, and extend RASL to the incremental gradient optimization framework. Our approach is faster than RASL and more robust to alignment than GRASTA. We can effectively and computationally efficiently learn the low-rank subspace from misaligned images, which is very practical for computer vision applications.

\subsection{Future Work}
Though this work presents an approach for robust image alignment more computationally efficient than state-of-the-art, a foremost remaining problem is how to scale the proposed approach to a very large streaming dataset such as is typical in real-time video processing.  
The fully online t-GRASTA algorithm presented here is a first step towards a truly large-scale real-time algorithm, but several practical implementation questions remain, including online parameter selection and error performance cross-validation.
Another question of interest is regarding the estimation of $d^k$ for the subspace update. Though we fix the rank $d$ in this paper, estimating $d^k$ and switching between Grassmannians is a very interesting future direction.

While preparing the conference version of this work \citep{he2013fg}, we noticed an interesting alignment approach proposed in~\citep{li_alignment_PAMI}. Though the two approaches of ours and~\citep{li_alignment_PAMI} are both obtained via optimization over a manifold, they perform  alignment for very different scenarios. For example, the approach in \citep{li_alignment_PAMI} focuses on semantically meaningful videos or signals, and then it can successfully align the videos of the same object from different views; t-GRASTA manipulates the set of misaligned images or the video of unstable camera to robustly identify the low-rank subspace, and then it can align these images according to the subspace. An intriguing future direction would be to merge these two approaches.

A final direction of future work is toward applications which require more aggressive background tracking than is possible by a GRASTA-type algorithm. For example, if a camera is following an object around different parts of a single scene, even though the background may be quickly varying from frame to frame, the camera will get multiple shots of different pieces of the background. Therefore, it may be possible to still build a model for the entire background scene using low-dimensional modeling. Incorporating camera movement parameters and a dynamical model into GRASTA would be a natural way to solve this problem, merging classical adaptive filtering algorithms with modern manifold optimization.

\section{ACKNOWLEDGEMENTS}
This work of Jun He is supported by NSFC (61203273) and by Collegiate Natural Science Fund of Jiangsu Province (11KJB510009).  Laura Balzano would like to acknowledge 3M for generously supporting her Ph.D. studies.





\bibliographystyle{elsarticle-num}
\bibliography{fg2013}







\end{document}